\documentclass[conference]{IEEEtran}
\IEEEoverridecommandlockouts
\usepackage{cite}
\usepackage{amsmath,amssymb,amsfonts}
\usepackage{algorithmic}
\usepackage{graphicx}
\usepackage{textcomp}
\usepackage{xcolor}

\usepackage{xspace}
\usepackage{mathtools}
\usepackage{kotex}

\usepackage{amsmath,amsfonts}
\usepackage{tabularx}  

\def\invcircledast#1{%
  \mathbin{\vphantom{\circledast}\text{%
    \ooalign{\smash{\blackcircle}\cr
             \hidewidth\smash{\textcolor{white}{\bf \footnotesize $#1$}}\hidewidth\cr
            }%
  }}%
}
\newcommand{\blackcircle}{\raisebox{-.6ex}{\scalebox{2.30}{$\bullet$}}}

\newcommand{\Design}{$\mathsf{DiTTO}$\xspace}

\usepackage{amsthm}

\usepackage{multirow}
\usepackage{threeparttable} 

\renewcommand{\baselinestretch}{0.95}
\usepackage[a4paper, total={184mm,253mm}]{geometry}
\def\BibTeX{{\rm B\kern-.05em{\sc i\kern-.025em b}\kern-.08em
    T\kern-.1667em\lower.7ex\hbox{E}\kern-.125emX}}
\begin{document}

\title{\huge Late Breaking Results: A Diffusion-Based Framework for Configurable and Realistic Multi-Storage Trace Generation \vspace{-15pt}} 

\author{
Seohyun Kim$^{1}$, Junyoung Lee$^{1}$, Jongho Park$^{1}$ \\
Jinhyung Koo$^{2}$ Sungjin Lee$^{2}$ and Yeseong Kim$^{1}$\\
$^1$DGIST, $^2$POSTECH\\
{\small \{selenium, lolcy3205, psyractal1123, yeseongkim\}@dgist.ac.kr, \{jhk5361, sungkin.lee\}@postech.ac.kr}

\thanks{\scriptsize
This work was supported by Institute of Information \& communications Technology Planning \& Evaluation (IITP) grant funded by the Korean government (MSIT) (No.2022-0-00991, 1T-1C DRAM Array Based High-Bandwidth, Ultra-High Efficiency Processing-in-Memory Accelerator).
This work was also supported by the Technology Innovation Program (RS-2024-00445759, Development of Navigation Technology Utilizing Visual Information Based on Vision-Language Models for Understanding Dynamic Environments in Non-Learned Spaces) funded by the Ministry of Trade, Industry \& Energy (MOTIE, Korea)
}
\vspace{-3mm}
}

\maketitle

\begin{abstract}
We propose \Design, a novel diffusion-based framework for generating realistic, precisely configurable, and diverse multi-device storage traces. Leveraging advanced diffusion techniques, \Design enables the synthesis of high-fidelity continuous traces that capture temporal dynamics and inter-device dependencies with user-defined configurations. Our experimental results demonstrate that \Design can generate traces with high fidelity and diversity while aligning closely with guided configurations with only 8\% errors.
\end{abstract}



\section{Introduction}

Understanding and optimizing the performance of distributed storage systems require detailed workload analysis in various environments such as RAID arrays and CephFS. Realistic workload traces, which capture read and write operations across multiple devices, are crucial for identifying performance bottlenecks and evaluating optimization strategies such as caching policies and load balancing. However, collecting real-world traces is costly and often impractical due to high instrumentation overheads and privacy concerns. This lack of accessible traces hinders the development of data-driven storage optimizations and limits the evaluation of emerging techniques.

To mitigate this, synthetic trace generation tools such as SPEC Storage allows for controlled workload simulation. While useful, these tools rely on predefined templates and fail to capture the complex, evolving behavior of real-world workloads. Machine learning-based methods~\cite{paul2022machine} attempt to improve realism but are still constrained by excessive feature engineering by human experts to identify application-specific characteristics.

In this paper, we propose \Design (\underline{\textbf{Di}}ffusion-based \underline{\textbf{T}}race generation and \underline{\textbf{T}}emporal \underline{\textbf{O}}utpainting), a \textit{diffusion-based} generative framework for producing realistic and configurable multi-device storage traces. Unlike conventional generative approaches that rely on predefined patterns or static sampling, our framework provides \textit{precise, user-controlled}, and \textit{arbitrary-length} trace generation, enabling workloads to be synthesized according to quantifiable constraints such as read/write ratios, access burstiness, and temporal dependencies.

\Design transforms raw access logs, including timestamps, operation types, and device identifiers, into structured\ multi-channel representations, enabling diffusion models to capture complex temporal dependencies and cross-device correlations. We introduce a workload conditioning mechanism that explicitly embeds quantifiable user-defined parameters into the generative process, providing fine-grained control over trace characteristics. Unlike conventional text-based conditioning in image diffusion models~\cite{dalle2}, our approach ensures explicit and constrained output control. Additionally, we design a sparsity-aware training approach to ensure that long periods of inactivity and bursty access patterns are accurately modeled. Finally, by leveraging outpainting techniques, a generative technique that extends content while preserving contextual coherence, \Design produces traces of arbitrary length while maintaining contextual consistency. In this paper, we make the following contributions.

\noindent 1) We propose a \textbf{diffusion-based synthetic trace generation framework} that accurately captures multi-device workload patterns through structured representations. To the best of our knowledge, it is the \underline{first work} that utilizes the diffusion technique for the storage trace generation. 

\noindent 2) We present a novel \textbf{quantifiable workload conditioning mechanism} that ensures quantifiable control over key workload properties such as read/write ratios.

\noindent 3) We design \textbf{sparsity-aware training} and \textbf{outpainting-based trace generation} techniques to generate workloads with long inactive periods and extended trace durations.

\noindent 4) We show that \Design can generate high-fidelity storage traces that replicate real-world workload characteristics while preserving trace diversity with less than 8\% error.

\begin{figure}
    \centering
    \includegraphics[width=0.95\columnwidth]{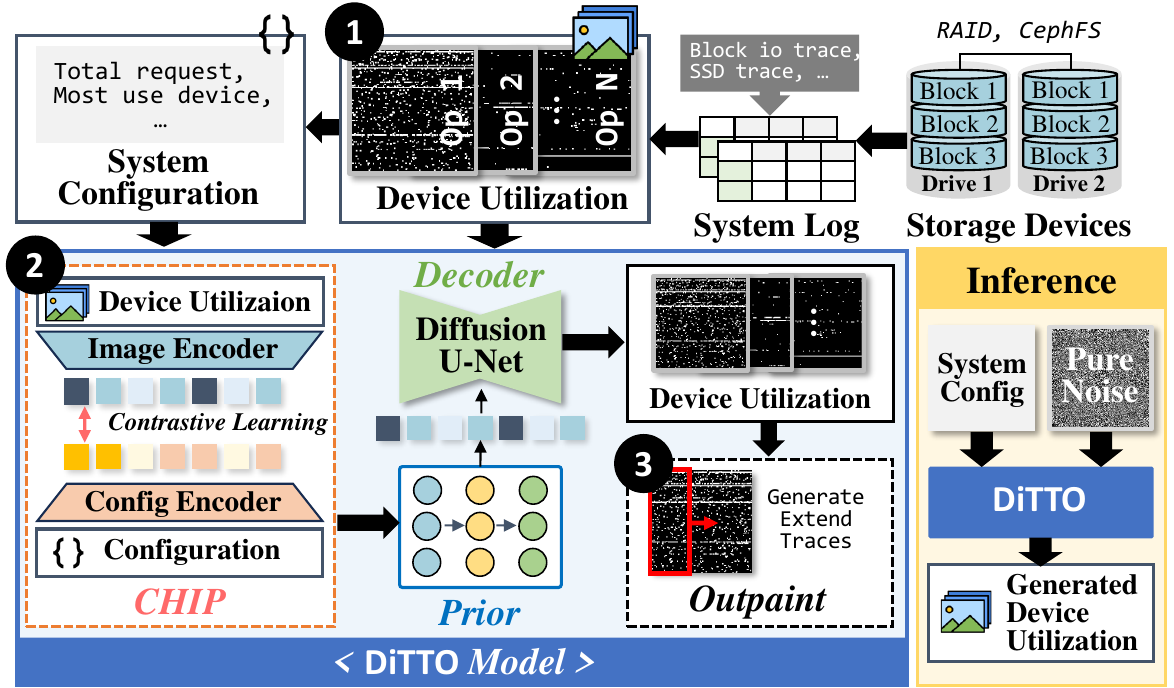}
    \caption{Overview of \Design}
    \label{fig:overview}
\end{figure}

\section{Proposed Technique: \Design}  
We propose \Design, a novel diffusion-based framework for generating realistic, configurable, and diverse multi-device storage traces. The framework captures intricate temporal dynamics and inter-device dependencies while allowing users to specify workload characteristics. The overall pipeline of \Design is illustrated in Fig.~\ref{fig:overview}, consisting of three main stages: $\invcircledast{1}$ transforming traces into structured image representations, $\invcircledast{2}$ encoding numeric workload characteristics for explicit conditioning, and $\invcircledast{3}$ generating arbitrarily long traces using a diffusion-based model enhanced with outpainting.

\noindent {\bf Transforming system traces into image-based representations.}  
\Design first converts raw per-device access logs into structured multi-channel image representations, enabling the diffusion model to learn workload patterns effectively. Each input trace consists of timestamped read and write operations mapped to corresponding storage devices. To construct the image representation, we segment time into fixed intervals along the x-axis, while the y-axis corresponds to different storage devices. Each pixel encodes the presence of a read or write event at a specific time interval and device, with separate channels distinguishing operation types.  
Storage workloads exhibit extreme sparsity, where access events are interspersed with long periods of inactivity. To mitigate this, we introduce local Gaussian noise augmentation, which creates smooth intensity gradients around access events. This prevents excessive sparsity in the image representation, improving the diffusion model to learn contextual relationships across time and devices.  

\noindent {\bf Encoding numeric workload characteristics.}  
A key requirement for synthetic trace generation is the ability to precisely control workload properties including read/write ratios, request rates, and device utilization patterns. To achieve this, we introduce CHIP (Contrastive Hyperconfiguration-Image Pretraining), a novel embedding mechanism that conditions the model on user-defined workload parameters.  
CHIP operates by mapping \textit{numeric workload configurations} into a latent space that aligns with the image-based representations of system traces. This is achieved via contrastive learning~\cite{clip}, where the model learns to associate similar workload characteristics with similar trace representations. Given a workload configuration, CHIP generates an embedding that guides the diffusion process, ensuring that generated traces adhere to the specified workload properties. Unlike conventional text-based conditioning in image diffusion models, CHIP explicitly enforces \textit{quantifiable constraints}, making it well-suited for structured data synthesis.  

\noindent {\bf Generating continuous traces.}  
Since real-world storage workloads span arbitrary durations, \Design must generate long sequences while preserving workload characteristics. To achieve this, we incorporate outpainting, a generative technique that extends trace durations while maintaining contextual coherence.  
The outpainting works by conditioning the diffusion model on the final states of previously generated traces, allowing seamless extension over time. Specifically, when generating a new trace segment, the model samples latent variables from the boundary of the prior output, ensuring that patterns such as periodic bursts, coordinated device accesses, and workload fluctuations remain consistent. This prevents abrupt transitions and maintains statistical fidelity across extended traces.

\ifx{
\section{Proposed Technique: \Design}
In this paper, we present \Design, a novel diffusion-based framework for generating realistic, configurable, and diverse multi-device storage traces. The framework captures the intricate temporal dynamics and inter-device dependencies characteristic of real-world workloads while supporting user-defined configurations for tailored trace generation. By leveraging a diffusion-based modeling approach inspired by DALL·E 2\cite{dalle2}, \Design synthesizes high-fidelity storage traces as structured data representations, meeting the requirements of diverse evaluation scenarios.

The training begins with transforming raw per-device access logs collected on a multi-storage environment into image representations, where each channel in the image corresponds to the operation type (read or write) and includes which storage devices are accessed within a fixed time window. This structured format enables the model to capture temporal patterns and inter-device correlations. User-defined configurations, including read/write ratios, device utilization patterns, and total requests, are embedded through our proposed embedding model, called CHIP (Contrastive Hyperconfiguration-Image Pretraining), which converts these annotations to the image-based workload representations in an embedding space. The CHIP embedding conditions the diffusion process via a prior, ensuring that generated traces reflect specified configurations. A UNet-based decoder then generates synthetic images from the latent representations, producing snapshots of system traces for fixed durations. For longer time horizons, \Design employs an outpainting technique to extend traces while maintaining temporal consistency.

\noindent {\bf Transforming system traces into image-based representations} Storage traces are characterized by infrequent events interspersed with long periods of inactivity, making direct encoding ineffective for generative modeling. To address this, \Design augments the image-like representation by introducing Gaussian noise around each access point. In this representation, the x-axis corresponds to time intervals within a fixed window, and each row along the y-axis is associated with a specific storage device. Each access event is encoded as a white pixel within the respective channel. 
Gaussian noise is applied locally around pixels to create a smooth intensity gradient, mitigating sparsity and enhancing the diffusion model's ability to capture contextual patterns around access events while maintaining spatial coherence.

\noindent {\bf Encoding numeric workload characteristics} To control the trace generation with fine-grained conditioning, e.g., read/write ratios or total requests, we devise the CHIP model, which maps numeric configurations into embedding vectors compatible with the image-based trace representations. CHIP aligns the high-dimensional numeric inputs with the latent space of the diffusion model using a contrastive learning approach. This alignment ensures that user-defined configurations, including quantifiable parameters like workload intensity or device utilization patterns, are accurately encoded as conditioning inputs.

\noindent {\bf Generating continuous trace} To extend generated traces beyond a fixed time window, \Design employes an outpainting technique that sequentially extends the trace while maintaining temporal and spatial consistency. Outpainting involves generating new trace segments by conditioning the model on the boundary of the previously generated trace. This approach preserves the contextual relationships across time intervals and ensures that patterns such as periodic bursts or coordinated device access are carried forward seamlessly. 
}\fi

\section{Experimental Results}

\begin{figure}
    \centering
    \includegraphics[width=1\columnwidth]{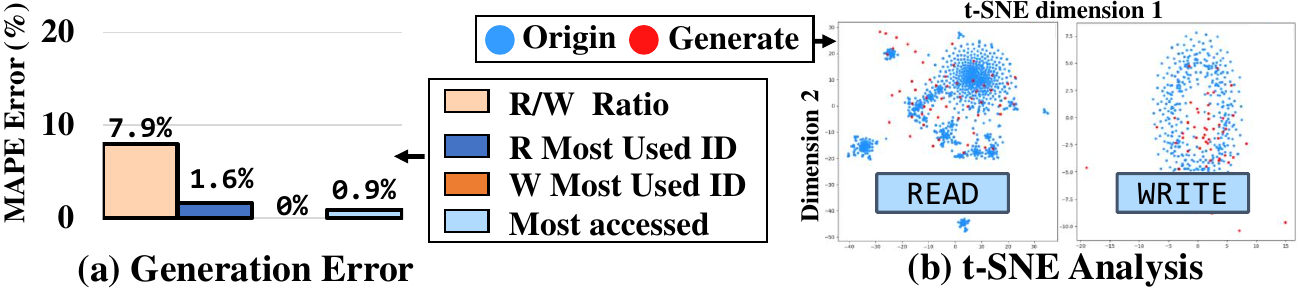}
    \caption{Generation error and t-SNE}
    \label{fig:eval}
\end{figure}

\begin{figure}
    \centering
    \includegraphics[width=1\columnwidth]{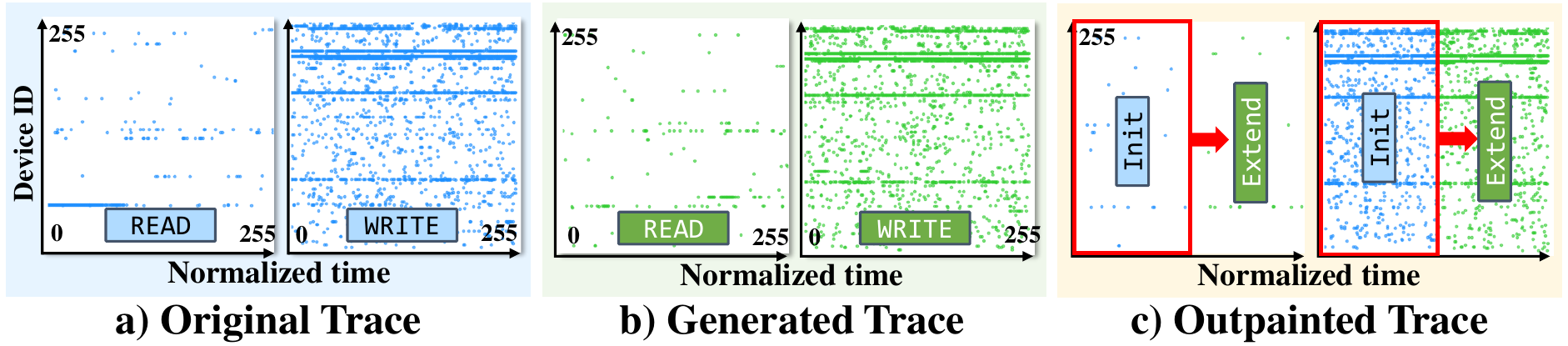}
    \caption{(a), (b) Comparison between the original and generated traces, (c) Extended trace with collected data and configuration (Outpainting)}
    \label{fig:img}
\end{figure}

\begin{figure}
    \centering
    \includegraphics[width=1\columnwidth]{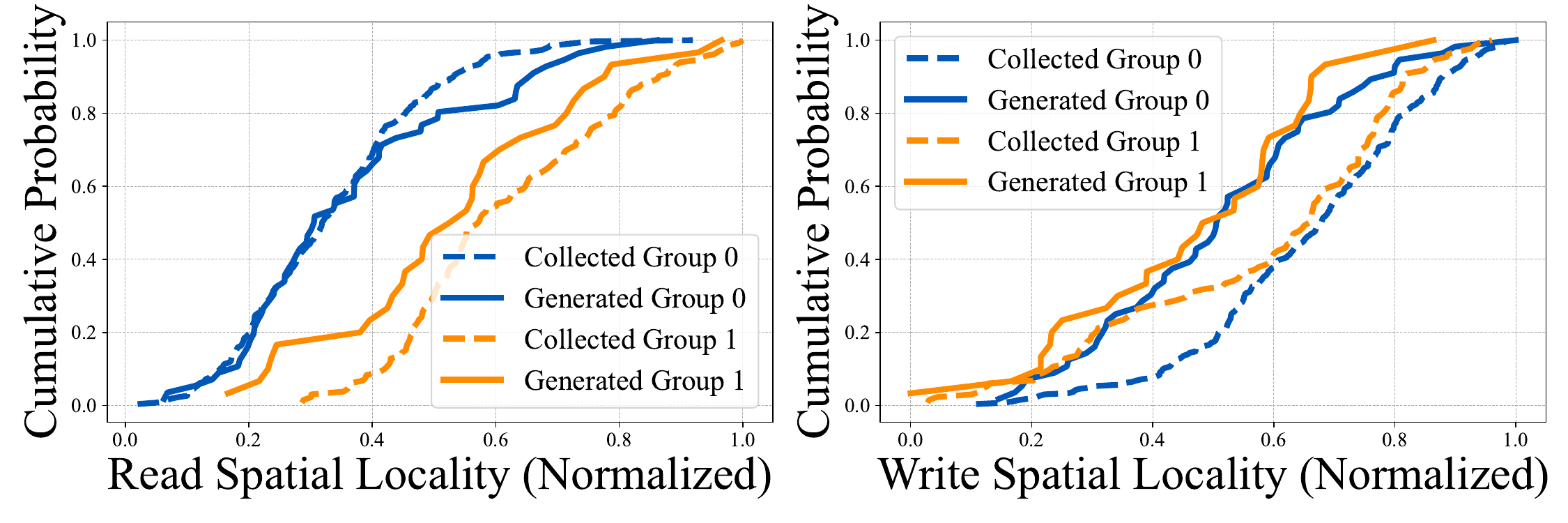}
    \caption{Distribution of spatial locality for different configuration groups}
    \label{fig:distribution}
\end{figure}

We implemented \Design using Pytorch 2.4 and evaluated the efficacy using Alibaba Block IO trace dataset.
One of the key features of \Design is its ability to generate realistic traces that closely adhere to user-specified configurations, such as read/write ratios and workload intensity. Figure~\ref{fig:eval}(a) presents the evaluation results for 50 generated traces, reporting the errors in how well the traces follow the user configurations. The results show that \Design achieves an average error of less than 8\% for read/write ratios and 2\% for device utilization patterns, demonstrating its precision in capturing workload characteristics. Additionally, the t-SNE analysis in Figure~\ref{fig:eval}(b) indicates that the generated traces exhibit trends similar to real (\textsf{Origin}) traces while maintaining sufficient diversity. These results confirm that the CHIP model effectively embeds quantifiable characteristics into the trace generation process. For instance, when tasked with generating traces for a read-heavy workload (82.99\% reads), \Design produces traces with a read ratio of 81.24\%, closely aligning with the intended configuration.  

Figure~\ref{fig:img} compares synthetic traces to their real-world counterparts to illustrate the fidelity of \Design. The results show that the generated traces (b) closely resemble real-world traces (a) under the same workload configuration, capturing realistic access patterns. Figure~\ref{fig:img}(c) demonstrates the results of the outpainting technique. The results present that \Design extends traces over a longer time horizon while preserving the consistency of previously observed patterns.

To further demonstrate how \Design generates distinguishable traces under different conditions, we clustered the collected traces into two groups and then generated traces by sampling the corresponding configurations of each cluster. Figure~\ref{fig:distribution} presents the distribution of real and generated traces in terms of read/write spatial locality. The results indicate that \Design successfully generates distinct traces that align with the patterns observed in each cluster, despite spatial locality \textit{not being explicitly included} in the user-configurable parameters. This suggests that \Design not only follows explicit workload parameters but also learns and preserves \textit{underlying high-level workload characteristics}, such as spatial locality, by capturing underlying access patterns from real traces.  

\section{Conclusion}
We propose \Design, which generates synthetic traces for distributed storage systems by introducing a diffusion-based framework. By integrating sparsity-aware training, CHIP for fine-grained conditioning, and outpainting for extended durations, \Design effectively captures intricate temporal and spatial patterns of real workloads. Our evaluation shows that \Design can generate realistic traces aligned with user configurations with the error of less than 8\%.

\bibliographystyle{short}
\bibliography{refer}


\ifx{
In this paper, we present \Design, a novel diffusion-based framework for generating realistic, configurable, and diverse multi-device storage traces. The proposed framework captures the intricate temporal dynamics and inter-device dependencies observed in real-world workloads while enabling user-defined configurations for tailored trace generation. As shown in Figure~\ref{}, the framework leverages a diffusion-based modeling learning technique, in particular, inspired by the architecture of DALL·E 2, to synthesize high-fidelity storage traces as structured data representations that meet the needs of diverse use scenarios.

The training process for the \Design diffusion model begins with the transformation of raw disk logs into multi-channel image representations. Each channel in the image corresponds to the operation type (read or write) and includes which storage devices are accessed within a fixed time window. This structured format allows the diffusion model to capture workload characteristics such as temporal patterns and inter-device correlations in a spatially organized manner. Additionally, each generated image is annotated with quantifiable characteristics derived from the raw system logs and transformed image-like data, such as read/write ratios, most used devices, and total requests. These annotations serve as user-configurable parameters, enabling the model to condition the trace generation process on specific workload configurations. 

At its core, the \Design diffusion model comprises three key components: the CHIP (Contrastive Hyperconfiguration-Image Pretraining) model, the prior, and the decoder. The CHIP model generates an embedding representation that integrates user-defined configurations into the trace generation process. Unlike conventional conditioning mechanisms, CHIP aligns configuration quantifiable annotations with image-based workload representations, ensuring compatibility and interpretability. This embedding is then processed by the prior model, which conditions the diffusion process to reflect the specified configurations. Finally, the UNet-based decoder reconstructs synthetic trace images from the conditioned latent representations, producing high-fidelity images that represent system trace snapshots for fixed time windows.
Once the model is trained, the \Design framework is capable of generating storage traces for fixed-duration time windows. To support longer time horizons, we integrate an outpainting technique that sequentially extends the generated traces beyond the initial window while maintaining temporal consistency.
}\fi

\ifx{
\section{Architecture}
asdfasfsfsdf

\begin{figure} [htbp]
    \centering
    \includegraphics[width=1\columnwidth]{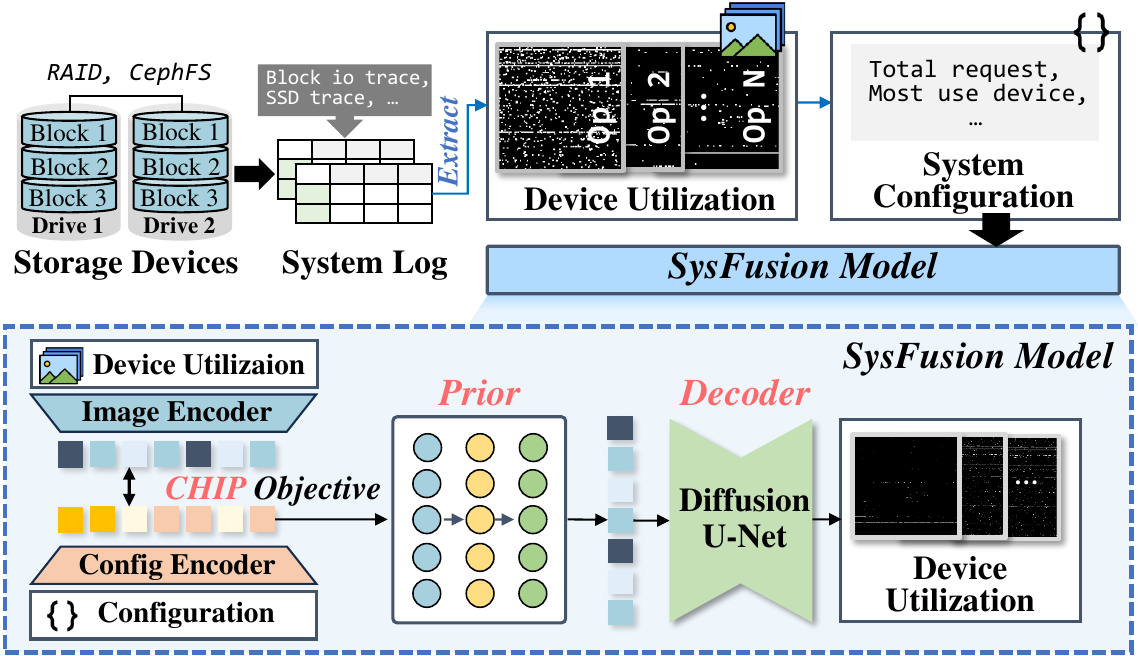}
    \caption{Overview of \Design}
    \label{fig:sysfusion}
\end{figure}

\subsection{Labeling System configuration}

\subsection{Representing System configuration and Images}
we present CHIP(Contrastive Hyperconfiguration-Image Pretraining), a novel neural network designed to efficiently learn visual concepts from system configurations.
Unlike the standard CLIP\cite{clip} architecture used in models like DALL·E 2\cite{dalle2}, which links images with textual descriptions, CHIP employs system configurations instead of textual inputs.

\subsection{Diffusion-based Image Generations}

\section{Evaluation}
asdfdsf

\begin{table}[htbp]
\caption{Table Type Styles}
\begin{center}
\begin{tabular}{|c|c|c|c|}
\hline
\textbf{Table}&\multicolumn{3}{|c|}{\textbf{Table Column Head}} \\
\cline{2-4} 
\textbf{Head} & \textbf{\textit{Table column subhead}}& \textbf{\textit{Subhead}}& \textbf{\textit{Subhead}} \\
\hline
copy& More table copy$^{\mathrm{a}}$& &  \\
\hline
\multicolumn{4}{l}{$^{\mathrm{a}}$Sample of a Table footnote.}
\end{tabular}
\label{tab1}
\end{center}
\end{table}

}\fi

\ifx{
\section*{Proposed Technique}
The proposed framework is designed to generate synthetic workload traces that represent the dynamic behavior of multi-device storage systems. Each trace consists of a sequence of access events, characterized by a timestamp, operation type (read or write), and the target storage device. To effectively capture and model these characteristics, the framework employs a three-step process: data encoding, diffusion-based generation, and decoding for trace reconstruction.

Temporal access logs are encoded into structured, multi-channel representations that preserve essential workload features while enabling the application of advanced generative techniques. In this representation, time is discretized into fixed intervals, forming the rows of a two-dimensional grid. Separate channels correspond to read and write operations, where the intensity at each grid cell reflects the frequency of accesses to a specific storage device during the corresponding time interval. This encoding scheme highlights temporal patterns, such as burstiness and idle periods, while maintaining the operational distinctions necessary for modeling.

The core of the framework is a diffusion model that learns to generate these structured representations. The model follows an iterative denoising process, where Gaussian noise is progressively added to the encoded data during the forward process, and the reverse process trains the model to reconstruct realistic data distributions. A sparsity-aware training regime ensures that the model focuses on meaningful patterns while avoiding overfitting to noise, a critical factor given the sparse nature of workload traces. By leveraging masked training and computationally efficient sparse convolutional layers, the model captures complex dependencies within the temporal and operational data.

Once the diffusion model generates synthetic representations, the decoding process transforms them back into trace logs. Timestamps are reconstructed based on the temporal discretization scheme, and operational patterns are extracted from the corresponding channels. Post-processing aligns the generated traces with user-defined configurations, allowing fine-grained control over characteristics such as read/write ratios, temporal locality, and access intensity distributions. This flexibility enables the framework to simulate a broad range of workloads, from read-heavy analytics operations to write-intensive database transactions.

\section*{Experimental Results}
The framework’s capabilities are evaluated through extensive experiments, focusing on two key aspects: adherence to user-defined configurations and the practical diversity of generated traces. Statistical analyses and visual inspections are conducted to validate the realism and utility of the synthetic traces.

To assess configurability, the framework is tested under scenarios with varying operational balances, including read-heavy, write-heavy, and balanced workloads. The results show that the generated traces closely follow the specified configurations, with minimal deviation in read/write ratios and temporal activity patterns. For example, in a scenario configured to produce a read-to-writratio of 3:1, the generated traces achieved a ratio of 2.98:1, demonstrating high fidelity to user-defined parameters. Temporal patterns, such as periodic bursts and idle intervals, were accurately replicated, as confirmed by correlation analyses with real-world traces.

Visualizations of the generated traces further illustrate their applicability. Figure 1 presents examples of synthetic traces for different configurations. Each visualization highlights the temporal and operational dynamics, showcasing the framework’s ability to model diverse workload behaviors. For instance, in a read-heavy scenario, the read operation channel exhibits dense activity clusters, while the write channel remains sparsely populated. In contrast, write-heavy traces show concentrated bursts in the write channel, demonstrating the adaptability of the proposed approach.

The diversity of generated traces is quantified using a normalized metric that evaluates coverage across a predefined configuration space. The proposed method achieves a score of 0.91, indicating robust capability to simulate varied workload scenarios. This diversity is essential for comprehensive storage system testing, enabling performance evaluations across multiple operational contexts.

While the primary evaluation focuses on synthetic trace realism and diversity, the potential applications of the framework extend further. By enabling precise control over workload characteristics, the proposed method provides a valuable tool for testing and optimizing storage system components. For example, in a use case targeting cache behavior, traces with high temporal locality can stress-test caching mechanisms, while traces with uniform distributions can evaluate system throughput under steady-state conditions. These applications, while not explicitly detailed due to space constraints, underscore the practical relevance of the proposed approach.

In summary, the experimental results validate the framework’s ability to generate synthetic traces that are not only realistic but also configurable and diverse. These attributes make it a powerful tool for advancing storage system design and evaluation.
}\fi

\ifx{
Understanding and optimizing the performance of distributed storage systems requires detailed workload analysis in environments like RAID arrays and CephFS. Realistic workload traces, which record read and write operations across multiple devices, allow identifying performance bottlenecks and enabling effective evaluation of optimization strategies, such as caching algorithms and load balancing. However, collecting real-world traces poses significant challenges due to high instrumentation costs, coupled with privacy concerns over sensitive data.
The lack of accessible traces impedes the development of efficient storage solutions, as enterprises are reluctant to publicly release sensitive data. 
As an alternative to real-world data collection, researchers have used tools capable of generating synthetic traces, e.g., Filebench and SPEC Storage. However, these tools are inherently static, failing to capture the rapidly evolving nature of storage environments; rather focusing narrowly on specific applications or devices.

In this paper, we claim that recent advancements in generative modeling, particularly diffusion models, offer a promising solution to the challenges of synthetic trace generation. Originally developed for image generation from user-guided text prompts, diffusion models are capable of learning complex data distributions through iterative denoising. This capability makes them well-suited for capturing the intricate dependencies observed in multi-device storage workloads. Unlike traditional synthetic trace generation tools, diffusion models can effectively represent sparsity, burstiness, and inter-device correlations. 
However, applying diffusion models to storage trace generation introduces unique challenges. First, storage traces exhibit inherent \textit{sparsity and high dimensionality}, with long periods of inactivity punctuated by bursts of activity and complex interdependencies across devices, necessitating specialized \underline{\textit{sparsity-aware training}} techniques. Second, practical utility requires that generated traces be configurable in a \textit{precise and quantifiable} manner, allowing \underline{\textit{flexible control over workload parameters}} such as read/write ratios and access burstiness to suit specific evaluation scenarios. Finally, typical diffusion models operate on fixed-size images, limiting their ability to produce traces of \underline{\textit{arbitrary length}}, which is a key requirement for evaluating long-term behaviors in storage systems. 

To address the limitations of existing synthetic trace generation tools, we propose a novel framework leveraging diffusion models to generate realistic and configurable multi-device storage traces. This framework transforms raw access logs, including timestamps, operation types (read/write), and device identifiers, into structured, image-like multi-channel representations. In particular, we focus on multi-storage device traces, which capture the timing of each device access.
By training the diffusion model on these representations, it learns the temporal dynamics and inter-device dependencies, iteratively refining random noise into coherent trace patterns. This process enables the model to accurately capture key workload characteristics, such as sparsity, burstiness, and cross-device correlations.

Our framework addresses storage trace generation challenges using diffusion models through sparsity-aware training that focuses on meaningful regions. It incorporates a workload conditioning model to embed user-defined configurations like read/write ratios and device utilization, enabling realistic and customizable trace generation. 
By leveraging outpainting~\cite{repaint}, it extends trace time horizons while preserving workload contexts, allowing for the generation of long-duration traces. Experimental results show \Design achieves less than 8\% error in replicating statistical properties and temporal patterns, while maintaining sufficient diversity.
}\fi

\end{document}